%% file: lrec-coling2024-example.tex
\documentclass[10pt, a4paper]{article}

\usepackage[]{lrec-coling2024} 

\usepackage{times}
\usepackage{latexsym}
\usepackage{multirow} 
\usepackage{booktabs} 
\usepackage{amssymb,amsthm,dsfont,mathrsfs}
\usepackage{amsmath}
\usepackage{multirow}
\usepackage{graphicx} 
\usepackage{subfigure}
\usepackage{color}
\usepackage{colortbl}
\usepackage{microtype}

\usepackage[utf8]{inputenc}
\input{math_commands.tex}

\title{Subspace Defense: Discarding Adversarial Perturbations \\
by Learning a Subspace for Clean Signals}

\name{
Rui Zheng\textsuperscript{1}*\thanks{*  Equal contribution.}, 
Yuhao Zhou\textsuperscript{1}*,
Zhiheng Xi\textsuperscript{1},
\\ \large{\textbf{
Tao Gui\textsuperscript{2\dag}\thanks{\textsuperscript{†} Corresponding authors.},
Qi Zhang\textsuperscript{1\dag}, 
Xuanjing Huang\textsuperscript{31\dag}
}}}

\address{
School of Computer Science, Fudan University\textsuperscript{1} \\
Institute of Modern Languages and Linguistics, Fudan University\textsuperscript{2} \\
International Human Phenome Institutes (Shanghai)\textsuperscript{3}\\
\{rzheng20, tgui, qz, xjhuang\}@fudan.edu.cn, zhouyh21@m.fudan.edu.cn\\
}

\abstract{
Deep neural networks (DNNs) are notoriously vulnerable to adversarial attacks that place carefully crafted perturbations on normal examples to fool DNNs.
To better understand such attacks, a characterization of the features carried by adversarial examples is needed.
In this paper, we tackle this challenge by inspecting the subspaces of sample features through spectral analysis. 
We first empirically show that the features of either clean signals or adversarial perturbations are redundant and span in low-dimensional linear subspaces respectively with minimal overlap, and the  classical low-dimensional subspace projection can suppress perturbation features out of the subspace of clean signals. 
This makes it possible for DNNs to learn a subspace where only features of clean signals exist while those of perturbations are discarded, which can facilitate the distinction of adversarial examples.
To prevent the residual perturbations 
that is inevitable in subspace learning, we propose an independence criterion to disentangle clean signals from perturbations.
Experimental results show that the proposed strategy enables the model to inherently suppress adversaries, which not only boosts model robustness 
but also motivates new directions of effective adversarial defense.
 \\ \newline  }

\begin{document}

\maketitleabstract

\section{Introduction}

Pre-trained language models (PLMs) are excellent feature extractors that map discrete inputs into fixed-length representations, which are then fed into a classifier for downstream tasks \cite{devlin-etal-2019-bert, DBLP:journals/corr/abs-1907-11692}.
Despite their great success, PLMs have been proven to be vulnerable to adversarial examples generated by placing perturbations on clean inputs \cite{ garg-ramakrishnan-2020-bae,DBLP:journals/tist/ZhangSAL20}.
Adversarial perturbations are often imperceptible to human, but can induce models to make erroneous predictions \cite{DBLP:journals/corr/GoodfellowSS14, ren-etal-2019-generating}.
Extensive researches have shown that adversarial vulnerabilities are prevalent in various NLP tasks, raising security issues in practical applications \cite{wallace-etal-2019-trick,zhang-etal-2021-crafting,lin-etal-2021-using, chen2023robust, zheng-etal-2023-detecting}.                                                   

Developing an understanding of the properties of adversarial perturbations and examples is a key requirement for adversarial defense.
\citet{DBLP:conf/iclr/TsiprasSETM19} and \citet{DBLP:conf/nips/IlyasSTETM19} argue that the non-robust parts of the features (those that generalize well but are brittle) can be manipulated by attackers to generate adversarial examples.
Moreover, the adversarial examples lie in low-probability data regions (not naturally occurring) and close to (but not on) clean data submanifold \cite{DBLP:journals/corr/SzegedyZSBEGF13, DBLP:journals/corr/TanayG16}.
\citet{DBLP:conf/cvpr/AkhtarLM18} further emphasize that adversarial perturbations on different data are highly correlated and redundant.
There exists a low-dimensional region, and perturbations belonging to this region can fool the model when added to any data point \cite{bao-etal-2023-casn}.
To summarize, the known properties of adversarial perturbations are: 1) they originate from non-robust features; 
2) they push data away from (but are close to) the clean data submanifold; and 3) they are highly correlated and redundant.

A number of methods are proposed to eliminate the affects of adversarial perturbations \cite{zheng-etal-2023-characterizing}.
Adversarial training is one of the most reliable techniques that generate adversarial examples for the training process and optimize model parameters to improve robustness \cite{DBLP:conf/iclr/MadryMSTV18, DBLP:conf/iclr/ZhuCGSGL20}.
\citet{DBLP:conf/focs/Allen-ZhuL21} point out that adversarial training works by purifying some dense and complex features, rather than removing non-robust features.
The information bottleneck-based methods \cite{DBLP:conf/iclr/AlemiFD017, DBLP:journals/entropy/FischerA20, DBLP:conf/nips/KimLR21} aim to filter out  excessive and noisy information that may invite adversarial attacks by compressing the mutual information between inputs and representations.
However, in high-dimensional feature spaces, the mutual information is calculated by approximate techniques with high cost.

In this paper, we show that, if studied from a feature perspective, a significant difference between clean signals and adversarial perturbations can be observed.
By applying principal component analysis (PCA) \cite{jolliffe2016principal} to the features encoded by PLMs, we show that the features of either clean signals or adversarial perturbations are redundant and lie in low-dimensional linear subspaces respectively with minimal overlap.
This suggests that adversarial perturbations can be suppressed by discarding features outside of the clean signal subspace.
To verify this, we project the adversarial examples onto clean subspaces, which significantly improves the robustness of the model while maintaining satisfactory performance on the main task.
Furthermore, as shown in Figure \ref{fig:projection}, we find that the clean subspace projector acts like a noise filter to eliminate the high feature magnitudes introduced by adversarial perturbations.

Based on the above analysis, we propose a new defense strategy, named subspace defense, which adaptively learns (with an auxiliary linear layer) a subspace for clean signals, where only features of clean signals exist while those of perturbations are discarded.
The subspace defense layer aims to learn a low-dimensional linear structure for the features of clean signals to retain as many task-relevant features as possible, and therefore inevitably preserve irrelevant features.
Therefore, we introduce the Hilbert-Schmidt independence criterion (HSIC) \cite{DBLP:conf/alt/GrettonBSS05} to ensure independence between preserved and discarded features, reducing the residual adversarial perturbations.
Our key contributions are summarized as follows:

\begin{itemize}
		\setlength{\itemindent}{0em}
		\setlength{\itemsep}{0em}
		\setlength{\topsep}{-0.5em}
	\item 
	We identify that, from a feature perspective, the features of either clean signals or adversarial perturbations are redundant and lie in low-dimensional linear subspaces respectively with minimal overlap, and the classical projection can suppress perturbation features outside the subspace of clean signals. 
		\item 
	We propose a new defense strategy, named subspace defense, which adaptively learns (with an auxiliary linear layer) a subspace for clean signals, where only features of clean signals exists while those of perturbations are discarded.
	\item 	
	We empirically show that our subspace defense strategy can consistently improve the robustness of PLMs. 
	Subspace defense strategy can accelerate the robustness convergence of adversarial training, thus avoid lengthy training processes.
\end{itemize}

\section{Related Work}
\subsection{Textual Adversarial Attack}

Text perturbation, unlike image attacks that operate in a continuous input space, needs to be performed discretly \cite{DBLP:journals/tist/ZhangSAL20, wang-etal-2021-textflint}.
Text attacks typically craft adversarial examples by deliberately manipulating characters \cite{ebrahimi-etal-2018-hotflip,DBLP:conf/sp/GaoLSQ18}, words \cite{ren-etal-2019-generating,DBLP:conf/aaai/JinJZS20, li-etal-2020-bert-attack, alzantot-etal-2018-generating, zang-etal-2020-word, DBLP:conf/aaai/MaheshwaryMP21}, phrases \cite{iyyer-etal-2018-adversarial}, or even the entire sentence \cite{wang-etal-2020-cat}. The word-level attacks, which are the most common ones, use the greedy algorithm \cite{ren-etal-2019-generating} or combinatorial optimization \cite{alzantot-etal-2018-generating} to search for the minimum number of substitutions.
Moreover, these attacks all guarantee the fluency of adversarial examples from the perspective of semantics \cite{li-etal-2020-bert-attack} or embedding space \cite{DBLP:conf/aaai/JinJZS20} to generate more stealthy adversarial examples.

\subsection{Textual Adversarial Defense}
In order to counter the adversarial attackers, a variety of defense methods are proposed to improve the robustness of the model \cite{DBLP:conf/acl/ZhengXLLGZHMSG23}.
Adversarial training-based approaches generate adversarial examples during the training process and optimize the model to defend against them \cite{DBLP:conf/iclr/MadryMSTV18, DBLP:conf/iclr/ZhuCGSGL20, liu-etal-2022-flooding}.
The information bottleneck approaches aim to filter out the redundant information contained in word embeddings and feature representations \cite{DBLP:conf/iclr/AlemiFD017, DBLP:journals/entropy/FischerA20, DBLP:conf/nips/KimLR21}.
\citet{zheng-etal-2022-robust} prove that it is possible to extract robust subnetworks from the pre-trained model, and these subnetworks can be used for robust training as a robust alternative to the original model \cite{xi-etal-2022-efficient1}.
In this paper, we eliminate non-robust and redundant features by projecting the perturbed features onto a low-dimensional subspace of clean signals.
Compared with the information bottleneck-based approaches, our method is more efficient and has a stronger suppression effect against adversarial noise.

\section{Spectral Analysis in Feature Space}

\begin{figure*}[t]
	\centering
	\subfigure[Spectral Analysis]{
		\centering
		\includegraphics[width=4.8cm]{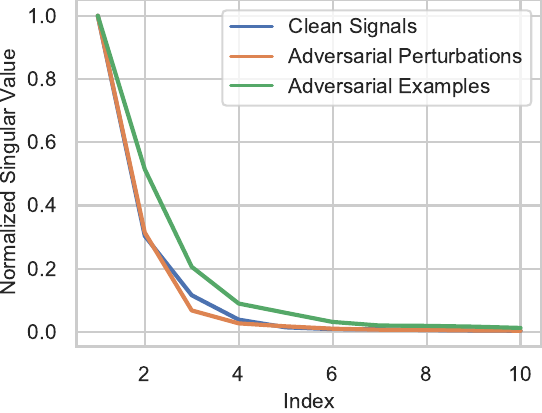}
		\label{fig:eig_values}
	}
	\hspace{0.2cm}
	\subfigure[Accuracy on Projected Features]{
		\centering
		\includegraphics[width=4.8cm]{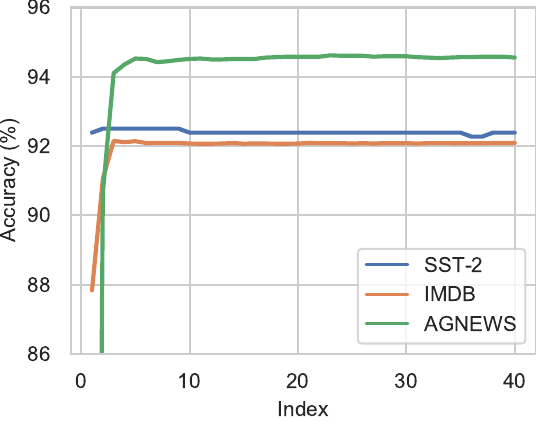}
		\label{fig:proj_acc}
	}
	\hspace{0.2cm}
	\subfigure[Robustness on Projected Features]{
		\centering
		\includegraphics[width=4.8cm]{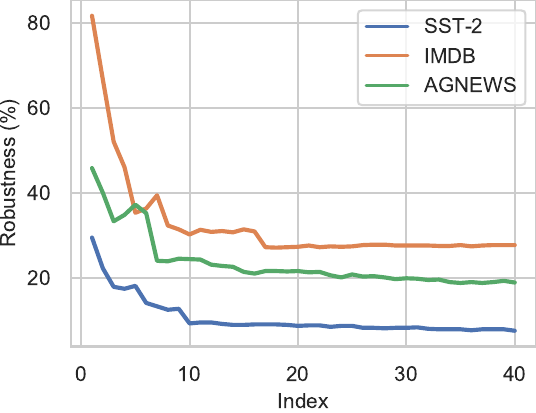}
		\label{fig:proj_aua}
	}

	\caption{(a) Spectral analysis of features of clean signals, adversarial perturbations, adversarial examples on SST-2.
	(b) and (c) respectively show the accuracy $(\%)$ and robustness evaluation (accuracy under TextFooler attack) after projecting the perturbed features on $p$-clean signal subspace.
	}
	\label{fig:spectral_analysis}
\end{figure*}

\begin{figure}[t]
	\centering
	\vspace{-0.2cm}
	\subfigure[Clean Signals]{
		\centering
		\includegraphics[width=6.7cm]{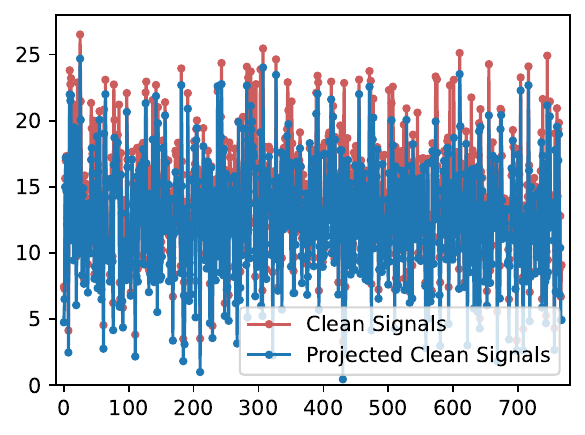}
		\label{fig:proj_clean}
	}
	\subfigure[Adversarial Examples]{
		\centering
		\includegraphics[width=6.7cm]{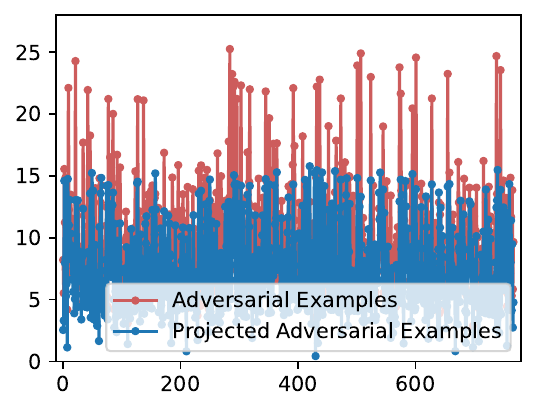}
		\label{fig:proj_adv}
	}
	\caption{
		Averaged feature magnitudes of clean signals, adversarial examples and their corresponding projected counterparts. 
		Low-dimensional $(p=2)$ clean subspace projector acts like a noise filter to eliminate the high feature magnitudes introduced by adversarial perturbations.
	}
	\label{fig:projection}
\end{figure}

In this section, we first show empirically that the features of either clean signals or adversarial perturbations are redundant and span in linear subspaces with minimal overlap between each other.
We then prove that the classical projection can suppress perturbation features outside the subspace of clean signals.

\subsection{Threat Model}

Given a dataset $\train=\{(\rvx_i,y_i)\}^N_{i=1}$ with $C$ classes, $\rvx_i$ denotes the \textbf{input embedding vector} and $y_i$ is the label.
Pre-trained language model $f(\cdot)$, such as BERT \cite{devlin-etal-2019-bert} and RoBERTa \cite{DBLP:journals/corr/abs-1907-11692}, is composed of a feature extractor $h(\cdot): \sR^{\train}\rightarrow \sR^d$ and a linear classifier $g(\cdot):\sR^d \rightarrow \sR^C$, where $d$ is the feature size.
In BERT-like models, the special \texttt{[CLS]} token at the top layer is used to aggregate contextual information and serve as feature representations.
Then features are fed to a classifier for sentence-level tasks, such as sentiment analysis or natural language inference.

\noindent
\textbf{Adversarial attack.} 
Consider a natural input $\rvx$ and a target threat model $f$, the adversarial attacker adds a small perturbation $\Delta \rvx$ to $\rvx$ such that: $\arg \max f(\rvx+\Delta \rvx) \neq y_\mathrm{true}$,
where $f(\rvx+\Delta \rvx)$ is the wrong predication of $\rvx+\Delta \rvx$, and $y_\mathrm{true}$ denotes the true class label of $\rvx$.
In practice, the perturbation $\Delta \rvx$ is often imperceptible to humans, and the perturbed input $\rvx'=\rvx+\Delta \rvx$ that causes the model to misclassify $\rvx$ is called as an adversarial example. 
In the NLP domain, $\Delta \rvx$ consists of adding, removing or replacing a set of phrases, words or characters in the original text.

\subsection{Spectral Analysis}

The learned feature of $\rvx
_i$ is $h(\rvx_i)$, after centralizing clean features (i.e. $\frac{1}{N}\sum_{i=1}^N h(\rvx_i)=0$), the clean feature matrix can be denoted as $\rmH=[h(\rvx_1), h(\rvx_2), \ldots, h(\rvx_N)]^T\in \sR^{N\times d}$ with $N \gg d$. 
We decompose the clean feature matrix $\rmH$ by singular value decomposition (SVD) \cite{horn2012matrix}:
\begin{equation}
	\rmH= \rmU \mathbf{\Sigma} \rmV^T,
\end{equation}
where $\rmU\in \sR^{N \times N}$ and $\rmV\in \sR^{d \times d}$ are orthogonal matrices, $\mathbf{\Sigma}$ is an $N \times d$ rectangular diagonal matrix with non-negative real numbers on the diagonal. 
The columns of $\rmU$ and $\rmV$ are called left and right singular vectors, respectively.
The diagonal entries $\sigma_i=\mathbf{\Sigma}_{i,i}$ of $\mathbf{\Sigma}$ are uniquely determined by $\rmH$ and are known as the singular values with $\sigma_1 \geq \sigma_2 \ldots \geq \sigma_n > 0$. The number of non-zero singular values is equal to the rank of $\rmH$. 
Similarly, the feature matrix of adversarial examples is $\rmH'=[h(\rvx'_1), h(\rvx'_2), \ldots, h(\rvx'_N)]^T\in \sR^{N\times d}$, and the feature matrix of perturbations is denoted as $\Delta \rmH = \rmH -\rmH'$.

\noindent
\textbf{Singular value distributions.}
We plot the singular values of the clean, adversarial and perturbation feature matrices to visualize their differences. 
All features are extracted from the test data of SST-2 and their corresponding adversarial examples.
As shown in Figure \ref{fig:eig_values}, the singular values of the natural and perturbation features decay faster than those of the adversarial features.
This means that the natural examples and the adversarial perturbations lie in low-dimensional feature subspaces, which are smaller than the feature subspace of adversarial examples.
More importantly, this indicates that there is little overlap between the clean feature subspace and the perturbation feature subspace.
These analyses provide the possibility of projecting adversarial examples onto the feature subspace of clean signals to suppress the adversarial perturbations \cite{DBLP:journals/spm/VaswaniBJN18}. 

\subsection{Subspace Projection}

Keeping the first $p$ principal components of the clean matrix $\rmH$, we use the first $p$ right singular vectors to generate a projection onto the subspace of clean signals for adversarial example $\rvx'_i$: 
\begin{equation}\label{subspace_projection}
	\hat{h}_p(\rvx'_i) = \rmV_p \rmV^T_p h(\rvx'_i),
\end{equation}
where $\rmV_p \in \sR^{d \times p}$ consists of the first $p$ right singular vectors, $\rmV_p \rmV^T_p \in \sR^{d \times d}$ denotes the orthoprojector onto the $p$-dimension subspace $\mathcal{M}_p$. 
This formulation can be interpreted as preserving the maximum feature component that can be characterized in $\mathcal{M}_p$.
We visualize the \textbf{accuracy} and \textbf{accuracy under attack} (for robust evaluation) of the projected adversarial features $\hat{h}_p(\rvx'_i)$.

\noindent
\textbf{Improving robustness.}
As shown in Figure \ref{fig:proj_acc}, the accuracy is high even in the low-dimensional subspace, which is consistent with the sharp decrease in singular values for accuracy.
In addition, the low-dimensional clean subspace projection can effectively improve the robustness of the model, while the adversarial subspace projection cannot.
Therefore, by performing clean subspace projection with appropriate dimensionality, not only the performance on  accuracy can be guaranteed, but also the effect of adversarial examples can be effectively reduced (although not completely avoided).

\noindent
Figure \ref{fig:projection} illustrates the averaged magnitudes of clean features, adversarial features, and their corresponding projected features.
The feature magnitudes of adversarial examples are generally higher than that of clean examples, and there are outliers in some dimensions that deviate significantly from the clean features.
Adversarial perturbation exhibits a significant distorting effect on the clean features, which leads to incorrect predictions of the classifier.
As shown in Figure \ref{fig:proj_adv}, the subspace projection can effectively narrow the magnitude gap between the features of clean and adversarial examples.

\section{Proposed Method}

Based on the above analysis, we introduce our subspace defense strategy, which dynamically learns a subspace in which only features of clean signals are preserved and features of perturbations are discarded.

\subsection{Subspace Learning Module}

Our previous results show that clean features span in a low-dimensional linear subspace.
When projecting learned features into this clean subspace, it is possible to obtain both good generalization and robustness.
Inspired by this, our goal is to learn an $r$-dimensional linear feature subspace for the clean examples and to remove redundant features that could be manipulated by the attacker.

In subspace learning module, we first apply a projection layer $\mathrm{Proj(\cdot)}$ to project the feature $h(\rvx_i)\in \sR^d$ onto the $r$-dimensional subspace $h(\rvx_i) \rightarrow \hat{h}_r(\rvx_i)\in \sR^r$, and then use the back-projection layer $\mathrm{Proj_b(\cdot)}$ to project $\hat{h}_r(\rvx_i)$ onto the original feature space $\hat{h}_r(\rvx_i) \rightarrow \hat{h}(\rvx_i)\in \sR^d$.
Formally, the final learned feature is:
\begin{equation}
	\begin{aligned}
		& \hat{h}_r(\rvx_i) = \mathrm{Proj}\left( h(\rvx_i)\right) , \\
		& \hat{h}(\rvx_i) = \mathrm{Proj_b}(\hat{h}_r(\rvx_i)),
	\end{aligned}
\end{equation}
where $\mathrm{Proj}(\cdot)$ and $\mathrm{Proj_b}(\cdot)$ are two linear layers defined as follows: $\mathrm{Proj}(h(\rvx_i))=\rmW_1 h(\rvx_i)+\rvb_1$ and $\mathrm{Proj_b}(\hat{h}_r(\rvx_i))=\rmW_2\hat{h}_r(\rvx_i)+\rvb_2$ with $\rmW_1 \in \sR^{r \times d}$, $\rmW_2 \in \sR^{d \times r}$, $\rvb_1 \in \sR^{r}$ and $\rvb_2 \in \sR^{d}$ are trainable parameters. 
Ideally, we would like to optimize the subspace learning module so that it maintains the underlying linear structure of the clean features.
Thus, we consider the reconstruction loss: \begin{equation}
	\gL_{recon} = \lVert h(\rvx_i) - \hat{h}(\rvx_i) \rVert^2_2.
\end{equation}
By using subspace learning module, features outside the clean subspace (typically adversarial perturbations) will be discarded.

\noindent
\textbf{Comparison with Autoencoder.}
The autoencoder operates by receiving data, compressing and encoding the data, and then reconstructing the data from the encoded representation \cite{kramer1991nonlinear, lecun2015deep}. The model is trained to minimize losses and the data is reconstructed as similar as possible. Through this process, the autoencoder learns important features of the data. 
The goal of both the autoencoder and our proposed method is to determine which aspects of the input need to be preserved and which can be discarded.
Our method differs from autoencoder in that: (1) We use linear layers as projectors, while autoencoder usually uses a nonlinear encoder and decoder; 
(2) An autoencoder obtains representations of samples from the input space, while our subspace learning module processes these representations to obtain a cleaner one.

\subsection{Hilbert-Schmidt Independence Criterion}

Subspace learning, such as PCA has the characteristic of being the optimal orthogonal transformation for keeping the subspace that has largest ``variance''.
In Equation  (\ref{subspace_projection}), the matrix $\rmV_p \rmV_p^T$ projects the features onto the $p$-dimensional subspace $\mathcal{M}_p$, while matrix $\mathbf{I}_p-\rmV_p \rmV_p^T$ denotes the projector onto the orthogonal complement space $\mathcal{M}_p^{\perp}$.
Therefore, preserved features and discarded features are independent of each other.
To achieve this property, we need to measure the degree of independence between the continues random variables $\hat{h}(\rvx_i)$ and $h(\rvx_i)-\hat{h}(\rvx_i)$ in high-dimensional spaces, and it is infeasible to rely on histogram-based measures. Thus, we chose to adopt the Hilbert-Schmidt Independence Criteria (HSIC) \cite{DBLP:conf/alt/GrettonBSS05}.

For two random variables $\rvu$ and $\rvv$, $\mathrm{HSIC}(\rvu,\rvv)$ is the Hilbert-Schmidt norm of the cross-covariance operator between $\rvu$ and $\rvv$ in Reproducing Kernel Hilbert Space (RKHS). HSIC is able to capture nonlinear dependencies between random variables, $\mathrm{HSIC}(\rvu,\rvv)=0$ if and only if $\rvu \perp \rvv$.
Formally, $\mathrm{HSIC}(\rvu,\rvv)$ is defined as:
\begin{equation*}
	\begin{aligned}
		\mathrm{HSIC}(\rvu,\rvv)  &= \E_{\rvu\rvv\rvu'\rvv'}[k(\rvu,\rvu')k(\rvv,\rvv')]\\
		+&\E_{\rvu\rvu'}[k(\rvu,\rvu')]\E_{\rvv\rvv'}[k(\rvv,\rvv')] \\
		-&2\E_{\rvu\rvv}[\E_{\rvu'}[k(\rvu,\rvu')]\E_{\rvv'}[k(\rvv,\rvv')]],
	\end{aligned}
\end{equation*}
where $\rvu'$, $\rvv'$ are independent copies of $\rvu$, $\rvv$, and $k$ is the radial basis function (RBF) kernel. 

In practice, the finite-sample estimates of $\mathrm{HSIC}(\rvu,\rvv)$ are used for statistical testing \cite{DBLP:conf/nips/GrettonFTSSS07}, feature similarity measures \cite{DBLP:conf/icml/Kornblith0LH19}, and model regularization \cite{DBLP:conf/cvpr/QuadriantoST19}.  
The unbiased estimator of $\mathrm{HSIC}(\rvu,\rvv)$ with $n$ samples can be defined as \cite{DBLP:journals/jmlr/SongSGBB12}:
\begin{equation*}
\begin{aligned}
	\mathrm{HSIC}(\rvu,&\rvv) =  \frac{1}{n(n-3)}\left[ tr(\tilde{\rvu} \tilde{\rvv}^T)\right. \\
	 &\left. +\frac{\bfone^T \tilde{\rvu} \bfone \bfone^T \tilde{\rvv}^T \bfone}{(n-1)(n-2)}-\frac{2}{n-2}\bfone^T\tilde{\rvu}\tilde{\rvv}^T\bfone\right],
\end{aligned}
\end{equation*}
where $\tilde{\rvu}_{ij}=(1-\delta_{ij})k(\rvu_i,\rvu_j)$ and $\tilde{\rvv}_{ij}=(1-\delta_{ij})k(\rvv_i,\rvv_j)$, i.e., the diagonal entries of $\tilde{\rvu}$ and $\tilde{\rvv}$ are set to zero.

To ensure the preserved features $\hat{h}(\rvx_i)$ and discard features $h(\rvx_i)-\hat{h}(\rvx_i)$ are independent of each other, we have:
\begin{equation}
	\gL_{hsic} = \sum_{i=1}^{n}  \mathrm{HSIC}\left( \hat{h}(\rvx_i),h(\rvx_i)-\hat{h}(\rvx_i)\right).
\end{equation}
Note that the reconstructed feature $\hat{h}(\rvx_i)$ is then fed to the classifier for downstream tasks.

\subsection{Model Training}

The subspace defense module can be used as an auxiliary component of the network and can be trained using standard training or different types of adversarial training.
Here, we take the original adversarial training as an example and define loss functions to train simultaneously the network and our subspace defense module:
\begin{equation} \label{final_loss}
	\gL_{all-at} = \gL_{adv} + \gL_{recon} + \lambda \cdot \gL_{hsic},
\end{equation}
where $\lambda>0$ is the balancing hyperparameter, $\gL_{adv}=\gL_{ce}(\rvx',\vtheta)$ is the adversarial loss, and the adversarial example $\rvx'$ is generated by:   
\begin{equation}
	\rvx' = \arg \max_{\lVert \rvx'-\rvx \rVert_F  \leq \epsilon} \gL(\rvx', y, \vtheta),
\end{equation}
where $\lVert \rvx'-\rvx \rVert_F  \leq \epsilon$ denotes the Frobenius normalization ball centered at $\rvx$ with radius $\epsilon$.

A wide range of attack methods have been proposed for the crafting of adversarial examples. 
For example, PGD iteratively perturbs normal example $\rvx$ for a number of steps $K$ with fixed step size $\eta$. 
If the perturbation goes beyond the $\epsilon$-ball, it is projected back to the $\epsilon$-ball \cite{DBLP:conf/iclr/MadryMSTV18}:
\begin{equation*}\label{eq:adv}
	\rvx_i^{k} = \prod\left(\rvx_i^{k-1}+
	\eta \cdot \mathrm{sign}(\nabla_\rvx \ell (\rvx_i^{k-1},y_i, \theta))\right),
\end{equation*}
where $\rvx_i^{k}$ is the adversarial example at the $k$-th step, $\mathrm{sign}(\cdot)$ denotes the sign function and $\prod(\cdot)$ is the projection function.

\section{Experiments}
\begin{table*}[t]
	\renewcommand\arraystretch{1.2}
	\setlength\tabcolsep{5pt}
	\centering
	\small
	\scalebox{1}{
	\begin{tabular}{llccc|cc|cc}
		\hline
		\hline
		\multicolumn{1}{l}{\multirow{2}{*}{\textbf{Dataset}}} &
		\multicolumn{1}{l}{\multirow{2}{*}{\textbf{Method}}} &
		\multicolumn{1}{c}{\multirow{2}{*}{\textbf{Clean$\%$}}} &
		\multicolumn{2}{c|}{\textbf{BERT-Attack}} & 
		\multicolumn{2}{c|}{\textbf{TextFooler}} & 
		\multicolumn{2}{c}{\textbf{TextBugger}} \\ 
		\cline{4-9}
		\multicolumn{1}{c}{} & \multicolumn{1}{c}{} & \multicolumn{1}{c}{}& 
		\multicolumn{1}{c}{\textbf{Aua$\%$}} &
		\multicolumn{1}{c|}{\textbf{\#Query}} & 
		\multicolumn{1}{c}{\textbf{Aua$\%$}} & 
		\multicolumn{1}{c|}{\textbf{\#Query}} & 
		\multicolumn{1}{c}{\textbf{Aua$\%$}} & 
		\multicolumn{1}{c}{\textbf{\#Query}} \\ 
		\cline{1-9}
		\hline
		\multirow{6}{*}{\textbf{IMDB}}
		& Fine-tune & $92.1$ & $4.0$ & $907.6$ & $5.3$ & $701.7$  & $15.0$ & $563.3$ \\ 
		& PGD  & $92.1$ & $9.0$ & $1886.5$ & $17.8$ & $1331.4$  & $36.2$ & $799.3$ \\ 
		&FreeLB & $92.0$ & $27.9$ & $2528.0$ & $34.7$ & $1724.7$ & $51.6$ & $1079.4$ \\
		&InfoBERT  & $92.1$ & $28.9$ & $2581.8$ & $36.0$ & $1744.9$ & $53.4$ & $1096.5$  \\
		&RobustT & $91.8$ & $69.9$ & $3262.2$ & $71.0$ & $2112.8$  & $71.9$ & $1139.9$ \\
		&\textbf{Ours} & $\mathbf{92.3}$ & $\mathbf{78.6}$ & $\mathbf{4275.6}$ & $\mathbf{78.6}$ & $\mathbf{2673.7}$  & $\mathbf{83.5}$ & $\mathbf{1749.5}$ \\
		\hline
		\multirow{6}{*}{\textbf{AGNews}}&
		Fine-tune & $94.7$ & $4.1$ & $412.9$ & $14.7$ & $306.4$  & $40.0$ & $166.2$ \\ 
		&PGD  & $\mathbf{95.0}$ & $20.9$ & $593.2$ & $36.0$ & $399.2$ & $56.4$ & $193.9$ \\
		&FreeLB  & $\mathbf{95.0}$ & $19.9$ & $581.8$ & $33.2$ & $396.0$ & $52.9$ & $201.1$ \\
		&InfoBERT  & $94.4$ & $11.1$ & $517.0$ & $25.1$ & $374.7$ & $47.9$ & $193.1$ \\
		&RobustT & $94.9$ & $21.8$ & $617.5$ & $35.2$ & $415.6$  & $49.0$ & $206.9$ \\
		&\textbf{Ours} & $93.8$ & $\mathbf{38.6}$ & $\mathbf{744.1}$ & $\mathbf{49.3}$ & $\mathbf{448.1}$ & $\mathbf{60.1}$ & $\mathbf{219.7}$ \\
		\hline
		\multirow{6}{*}{\textbf{SST-2}}&
		Fine-tune & $92.1$ & $3.8$ & $106.4$ & $6.1$ & $90.5$ & $28.7$ & $46.0$ \\ 
		&PGD & $\mathbf{92.2}$ & $13.4$ & $151.3$ & $18.1$ & $118.5$ & $44.2$ & $53.6$\\
		&FreeLB  & $91.7$ & $23.9$ & $174.7$ & $29.4$ & $132.6$ & $49.7$ & $53.8$ \\
		&InfoBERT  & $92.1$ & $14.4$ & $162.3$ & $18.3$ & $121.4$ & $40.3$ & $51.2$ \\
		&RobustT & $90.9$ & $20.8$ & $169.2$ & $28.6$ & $149.8$ & $43.1$ & $53.9$ \\
		&\textbf{Ours} & $91.3$ & $\mathbf{36.5}$ & $\mathbf{201.2}$ & $\mathbf{46.3}$ & $\mathbf{167.3}$  & $\mathbf{54.5}$ & $\mathbf{62.3}$ \\
		\hline
		\hline
	\end{tabular}
}
	\caption{Main results on adversarial robustness evaluation. 
	The proposed method achieves a significant improvement of robustness compared with other baselines.
	The best performance is marked in bold. 
}
	\label{tab:results_main}
\end{table*}

In this section, we conduct several experiments to demonstrate the effectiveness of our method over  multiple NLP tasks such as text classification. We first compare our method against five competitive baselines in terms of accuracy on clean datasets and robust evaluation. Then, we perform an ablation study to confirm the importance of adversarial loss objective.
We use widely adopted BERT$_{\mathrm{BASE}}$ as the backbone model which is implemented by Huggingface Transformers\footnote{https://github.com/huggingface/transformers} \citep{wolf-etal-2020-transformers} library.

\subsection{Datasets}

We consider three commonly used text classification datasets in our experiments: Stanford Sentiment Treebank of binary classification (SST-2) \citep{socher-etal-2013-recursive}, AG News corpus (AGNews) \citep{DBLP:conf/nips/ZhangZL15} and Internet Movie Database (IMDB) \citep{maas-etal-2011-learning}. The first two are binary sentiment analysis tasks that classify reviews into positive or negative sentiment, and the last is a classification task in which articles are categorized as world, sports, business or sci/tech.

\subsection{Baselines}
We compare our method against the basic fine-tuning method and five competitive adversarial defense methods in terms of accuracy on clean datasets and robust evaluation. (1) Fine-tune \citep{devlin-etal-2019-bert}: The official implementation for BERT on downstream tasks. (2) FreeLB \citep{DBLP:conf/iclr/ZhuCGSGL20}: An enhanced gradient-based adversarial training method which is not targeted at specific attack methods. (3) PGD \citep{DBLP:conf/iclr/MadryMSTV18}: Projected gradient descent that formulates adversarial training algorithms into solving a min-max problem that minimizes the empirical loss on adversarial examples  that can lead to maximized adversarial risk. (4) InfoBERT \citep{DBLP:conf/iclr/WangWCGJLL21}: A learning framework for robust fine-tuning of PLMs from an information-theoretic perspective. (5) RobustT \citep{zheng-etal-2022-robust}: Robust Lottery Ticket Hypothesis finds the full PLM contains subnetworks, i.e., robust tickets, that can achieve a better robustness performance.

\subsection{Attack Methods and Evaluation Metrics}

Three well-received attack methods are adopted to evaluate our method against baselines. (1) BERT-Attack \citep{DBLP:conf/emnlp/LiMGXQ20} generates adversarial samples using pre-trained masked language models exemplified by BERT, which can generate fluent and semantically preserved samples. (2) TextFooler \citep{DBLP:conf/aaai/JinJZS20} identifies the words in a sentence which is important to the victim model, and then replaces them with synonyms that are semantically similar and syntactically correct until the model's prediction for that sentence changes. (3) TextBugger \citep{DBLP:conf/ndss/LiJDLW19} generates misspelled words by using character-level and word-level perturbations. We use TextAttack\footnote{https://github.com/QData/TextAttack} toolkit to implement these attack methods in adversatial attack experiments.

The evaluation metrics used in our experiments are shown bellow: Clean accuracy (\textbf{Clean\%}) denotes the accuracy on the test dataset. Accuracy under attack (\textbf{Aua\%}) represents the accuracy under adversarial attacks. 
 Number of queries (\textbf{\#Query}) refers to the average number of queries made by the attacker to the victim model. For the same attack method, models with higher robustness are expected to have higher clean accuracy, accuracy under attack and number of queries in the robustness evaluation.

\begin{table}[tbp]
	\renewcommand\arraystretch{1.2}
	\setlength\tabcolsep{5pt}
	\centering
	\small
	\scalebox{0.91}{
	\begin{tabular}{clccc}
		\hline
		\hline
		\multicolumn{1}{c}{\multirow{2}{*}{\textbf{Dataset}}} &
		\multicolumn{1}{c}{\multirow{2}{*}{\textbf{Method}}} &
		\multicolumn{1}{c}{\multirow{2}{*}{\textbf{Clean$\%$}}} &
		\multicolumn{2}{c}{\textbf{Aua$\%$}}
		\\ \cline{4-5}
		\multicolumn{1}{c}{} & \multicolumn{1}{c}{} & \multicolumn{1}{c}{}& 
		\multicolumn{1}{c}{\textbf{\scriptsize TextFooler}} &
		\multicolumn{1}{c}{\textbf{\scriptsize TextBugger}}   \\ 
		\cline{1-4}
		\hline
		\multirow{6}{*}{\textbf{QNLI}}&
		Fine-tune & $\mathbf{91.6}$ & $4.7$  & $10.5$ \\
		&PGD  & $91.2$ & $12.2$ & $18.7$ \\
		&FreeLB  & $90.5$ & $12.8$ & $12.0$ \\
		&InfoBERT  & $91.5$ & $16.4$ & $20.9$ \\
		& RobustT & $91.5$ & $17.0$ & $25.9$  \\
		& Ours & $91.4$ & $\mathbf{33.5}$ & $\mathbf{43.1}$  \\
		\hline
		\multirow{6}{*}{\textbf{MNLI}}&
		Fine-tune & $\mathbf{84.4}$ & $7.7$  & $4.3$   \\ 
		&PGD  & $83.9$ & $14.5$ & $15.7$  \\
		&FreeLB  & $82.9$ & $11.0$ & $8.4$  \\
		&InfoBERT  & $84.1$ & $10.8$ & $8.4$  \\
		&RobustT & $84.0$ & $18.4$ & $22.6$   \\
		&Ours & $84.2$ & $\mathbf{21.6}$ & $\mathbf{33.5}$   \\
		\hline
		\multirow{6}{*}{\textbf{QQP}}&
		Fine-tune & $91.3$ & $24.8$  & $27.8$ \\ 
		&PGD & $91.2$ & $32.0$  & $33.5$  \\
		&FreeLB & $91.2$ & $27.4$  & $28.1$  \\
		&InfoBERT & $\mathbf{91.9}$ & $34.4$  & $35.9$  \\
		&RobustT & $91.5$ & $\mathbf{47.2}$  & $\mathbf{46.0}$ \\
		&Ours & $91.6$ & $34.1$  & $35.3$ \\
		\hline
		\hline
	\end{tabular}
}
	\caption{Adversarial robustness evaluation of the proposed method on QNLI, MNLI and QQP datasets.}
	\label{tab:glue}
\end{table}

\section{Results and Discussions}

In this section, we illustrate the effectiveness of our approach and show the impact of the individual components in our approach on the model's robustness.

\subsection{Main Results on Robustness Evaluation}
From the results shown in Table \ref{tab:results_main}, we can observe that: 1) 
The proposed approach achieves significant robustness improvements compared to other defense baselines.
This is because the proposed method can remove perturbations by projecting feature representations onto a subspace of clean signals without compromising task accuracy.
2) Although InfoBERT can also filter out redundant information by compressing the mutual information between inputs and representations, its performance is far from satisfactory.
This suggests that mutual information is difficult to optimize in a high-dimensional feature space and is not as effective as directly removing low-contributing features.

We also evaluate the performance of our proposed approach on other tasks, such as natural language inference and paraphrase identification.
As seen in Table \ref{tab:glue}, our proposed method consistently improves the robustness of the model on the QNLI, MNLI, and QQP datasets.

\subsection{Ablation Study}

\begin{table}[t]
	\renewcommand\arraystretch{1.4}
	\setlength\tabcolsep{5pt}
	\centering
	\small
	\scalebox{1.01}{
	\begin{tabular}{llcc}
		\hline
		\hline
		\textbf{Dataset}   & \multicolumn{1}{c}{\textbf{Method}} & \textbf{Clean$\%$} & \textbf{Aua$\%$} \\ \hline
		\multirow{4}{*}{\textbf{IMDB}}   & \textbf{Ours}               &$92.3$ & $\mathbf{78.6}$   \\
		& \quad \textbf{w/o} Projector   & $92.0$  & $34.7$   \\ 
		& \quad \textbf{w/o} HSIC   & $\mathbf{92.5}$  & $41.3$   \\ 
		& \quad \textbf{w/o} Adv                    & $92.2$     & $45.5$   \\ \hline
		\multirow{4}{*}{\textbf{AGNews}} & \textbf{Ours}              & $93.8$     & $\mathbf{49.3}$   \\ 
		& \quad \textbf{w/o} Projector   & $\mathbf{95.0}$  & $33.2$   \\ 
		& \quad \textbf{w/o} HSIC   & $93.9$  & $37.3$   \\ 
		& \quad \textbf{w/o} Adv                    & $93.5$     & $41.0$   \\ \hline
		\multirow{4}{*}{\textbf{SST-2}}   & \textbf{Ours}              & $91.3$     & $\mathbf{43.6}$   \\ 
		& \quad \textbf{w/o} Projector   & $91.7$  & $29.4$   \\ 
		& \quad \textbf{w/o} HSIC   & $\mathbf{92.3}$  & $26.5$   \\ 
		& \quad \textbf{w/o} Adv                    & $92.1$     & $36.4$  \\ \hline\hline
	\end{tabular}
}
	\caption{Ablation study on text classification datasets. \textbf{Aua$\%$} is obtained after using TextFooler attack.}
	\label{tab:ablation_main}
\end{table}

To illustrate the contribution of each component of our method, we perform the ablation study with the following components removed: the projector to clean signal subspace (projector), Hilbert-Schmidt independence criteria (HSIC), and adversarial examples (Adv).
We can observe that: 1) The subspace projector is important for performance and the improvements in our approach come mainly from this component.
2) Both independence criteria and adversarial training can further help our approach to improve robustness, which also indicates that our approach can be well integrated with existing adversarial training.

\subsection{Effect of Subspace Dimension}

\begin{figure}[t]
	\centering
	\vspace{-0.2cm}
	\subfigure[Accuracy]{
		\centering
		\includegraphics[width=6cm]{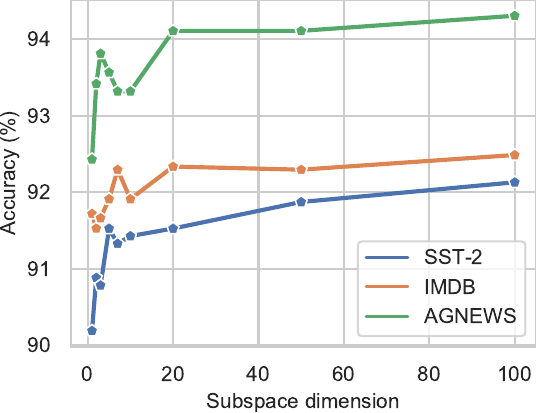}
		\label{fig:dimension_acc}
	}
	\subfigure[Robustness]{
		\centering
		\includegraphics[width=6cm]{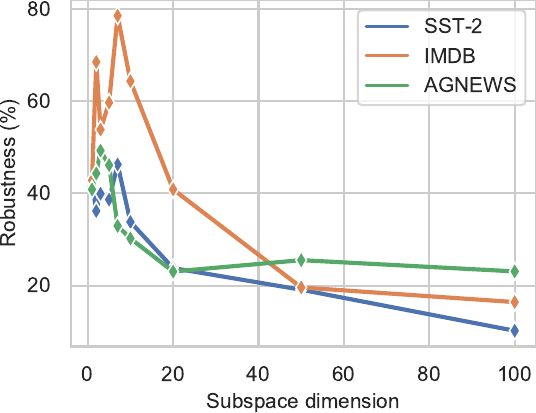}
		\label{fig:lineplot_dimension_aua}
	}
	\caption{Accuracy and robustness evaluation (accuracy under TextFooler attack) of model under different subspace dimension, both of which reach the peak when the subspace dimension is between 5 and 10.
	}
	\label{fig:dimension}
\end{figure}

In Figure \ref{fig:dimension}, we show the performance of our proposed method as the subspace dimension increases.
When the dimension increases to a certain level, the task accuracy reaches its peak, which indicates that we need a suitable dimension to achieve good task performance and larger dimensions are useless.
The robustness of the model can be maintained at a high level in low-dimensional subspaces until a certain threshold; beyond this threshold, the robustness deteriorates.
The dimension controls the extent to which features are discarded, and no perturbations are discarded when the dimension is too large.
When the subspace dimension is between $5$ and $10$, our proposed method can achieve a win-win result in terms of accuracy and robustness.

\subsection{Speedup Robust Training}
\begin{figure}[t]
	\centering
	\vspace{-0.2cm}
	\subfigure[SST-2]{
		\centering
		\includegraphics[width=6cm]{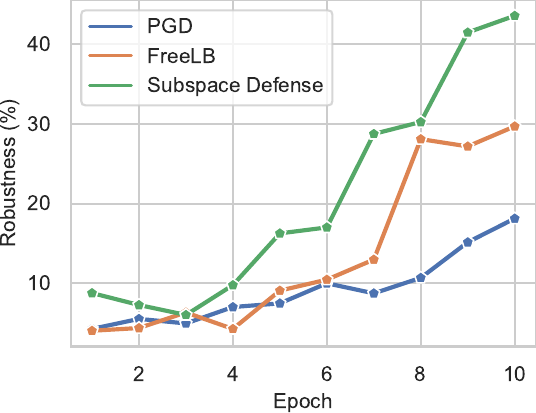}
		\label{fig:speedup_sst2}
	}
	\subfigure[AGNews]{
		\centering
		\includegraphics[width=6cm]{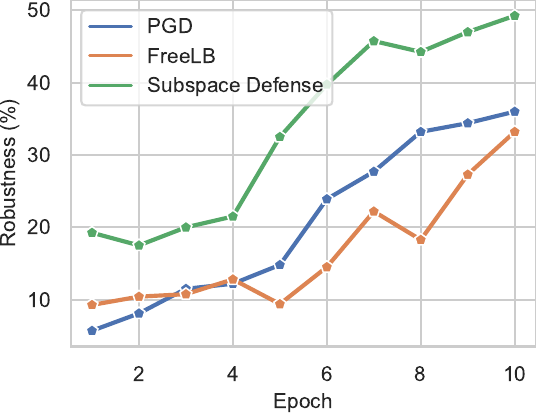}
		\label{fig:speedup_agnews}
	}
	\caption{Robustness of each epoch throughout training on SST-2 and AGNews with different training strategy. Compared to adversarial training methods like PGD and FreeLB, the proposed subspace defense speeds up robust training and converges much faster in terms of accuracy under attack.
	}
	\label{fig:speedup_robust}
\end{figure}

An important property of the proposed method is to accelerate the convergence of the training process.
The training curves in Figure \ref{fig:speedup_robust} show that the proposed method converges much faster in terms of robustness on the SST-2 and AGNEWS datasets compared to adversarial training methods such as PGD and FreeLB.
That is because learning the low-dimensional structure of clean signals is easier than learning to resist adversarial examples.
The advantage in convergence speed will make our method easy to apply in practice.

\subsection{Impact of Projector Structure}

\begin{table*}[t]
	\renewcommand\arraystretch{1.2}
	\setlength\tabcolsep{5pt}
	\centering
	\small
	\scalebox{1}{
	\begin{tabular}{lll|ll|ll|ll} 
		\hline
		\hline
		\multirow{2}{*}{Structures} & \multicolumn{2}{c|}{3 dimension} & \multicolumn{2}{c|}{5 dimension} & \multicolumn{2}{c|}{7~dimension} & \multicolumn{2}{c}{10~dimension}  \\
		\cline{2-9}
		& \textbf{Clean$\%$}  & \textbf{Aua$\%$}                       & \textbf{Clean$\%$}  & \textbf{Aua$\%$}                       & \textbf{Clean$\%$}  & \textbf{Aua$\%$}                       & \textbf{Clean$\%$}  & \textbf{Aua$\%$}                        \\ 
		\hline
		1-layer MLP                 & $91.3$ & $23.0$                      & $91.6$ & $24.1$                      & $91.5$ & $23.5$                      & $92.1$ & $19.9$                       \\
		1-layer MLP $+$ ReLU            & $91.9$ & $12.6$                      & $91.2$ & $23.5$                      & $91.2$ & $23.9$                      & $91.5$ & $17.1$                       \\
		2-layer MLP                 & $\mathbf{92.0}$ & $\mathbf{26.4}$                     & $\mathbf{92.1}$ & $\mathbf{31.1}$                      & $\mathbf{92.2}$  & $\mathbf{28.9}$                      & $\mathbf{92.2}$ & $\mathbf{28.8}$                       \\
		2-layer MLP+ReLU            & $\mathbf{92.0}$ & $17.1$                      & $\mathbf{92.1}$ & $27.7$                      & $91.9$ & $24.5$                      & $92.1$ & $18.4$                       \\
		\hline
		\hline
	\end{tabular}
}
	\caption{Impact of projector structure. Four structures are compared across different subspace dimensions on SST-2 in terms of accuracy and accuracy under attack. Structure with more layers performs slightly better and non-linearity impairs the robustness.
	}
	\label{tab:types}
\end{table*}

To better understand the impact of the structure of the autoencoder on performance, we compare the following structures: $1$-layer linear MLP, 1-layer MLP $+$ ReLU,  2-layer linear MLP (applied in our method), and 2-layer MLP $+$ ReLU.
Table \ref{tab:types} shows that all structures help to improve the robustness of the model.
More layers are slightly better for autoencoder, but non-linearity is harmful.
The non-linearity destroys the linear structure of the subspace, so that even subspaces in low dimensions may contain perturbations.
The above results show that the robustness of the model can be significantly improved by adding a simple autoencoder to the existing model.

\section{Conclusion}
In this paper, we characterize the feature properties of both clean signals and adversarial perturbations via low-dimensional subspace projection.
Further, we provide an initial intuition as to how subspace learning is an effective method for defending adversaries, which suggests that subspace projector eliminates feature magnitudes of adversarial perturbations. Further investigation in this direction may lead to new techniques for both adversarial attack and defense.

\section{Acknowledgements}
The authors wish to thank the anonymous reviewers for their helpful comments. This work was partially funded by National Natural Science Foundation of China (No. 62206057, 61976056, 62076069, 61906176), Shanghai Rising-Star Program (23QA1400200), Natural Science Foundation of Shanghai (23ZR1403500), Program of Shanghai Academic Research Leader under grant 22XD1401100, CCF-Baidu Open Fund, and CCF-Baichuan Fund.
The computations in this research were performed using the CFFF platform of Fudan University.



\section{Bibliographical References}\label{sec:reference}

\bibliographystyle{lrec-coling2024-natbib}

\bibliography{lrec-coling2024-example, anthology, custom}


\end{document}

%% file: math_commands.tex
\usepackage{amsmath,amsfonts,bm}

\def\bfone{{\mathbf{1}}}

\def\eqref#1{equation~\ref{#1}}

\def\1{\bm{1}}
\newcommand{\train}{\mathcal{D}}

\def\rvb{{\mathbf{b}}}

\def\rvu{{\mathbf{i}}}

\def\rvu{{\mathbf{u}}}
\def\rvv{{\mathbf{v}}}

\def\rvx{{\mathbf{x}}}

\def\rmH{{\mathbf{H}}}

\def\rmU{{\mathbf{U}}}
\def\rmV{{\mathbf{V}}}
\def\rmW{{\mathbf{W}}}

\def\vtheta{{\bm{\theta}}}

\DeclareMathAlphabet{\mathsfit}{\encodingdefault}{\sfdefault}{m}{sl}
\SetMathAlphabet{\mathsfit}{bold}{\encodingdefault}{\sfdefault}{bx}{n}

\def\gL{{\mathcal{L}}}

\def\sR{{\mathbb{R}}}

\newcommand{\E}{\mathbb{E}}



%% file: lrec-coling2024-example.bbl
\begin{thebibliography}{54}
\expandafter\ifx\csname natexlab\endcsname\relax\def\natexlab#1{#1}\fi

\bibitem[{Akhtar et~al.(2018)Akhtar, Liu, and Mian}]{DBLP:conf/cvpr/AkhtarLM18}
Naveed Akhtar, Jian Liu, and Ajmal Mian. 2018.
\newblock \href {https://doi.org/10.1109/CVPR.2018.00357} {Defense against
  universal adversarial perturbations}.
\newblock In \emph{2018 {IEEE} Conference on Computer Vision and Pattern
  Recognition, {CVPR} 2018, Salt Lake City, UT, USA, June 18-22, 2018}, pages
  3389--3398. Computer Vision Foundation / {IEEE} Computer Society.

\bibitem[{Alemi et~al.(2017)Alemi, Fischer, Dillon, and
  Murphy}]{DBLP:conf/iclr/AlemiFD017}
Alexander~A. Alemi, Ian Fischer, Joshua~V. Dillon, and Kevin Murphy. 2017.
\newblock \href {https://openreview.net/forum?id=HyxQzBceg} {Deep variational
  information bottleneck}.
\newblock In \emph{5th International Conference on Learning Representations,
  {ICLR} 2017, Toulon, France, April 24-26, 2017, Conference Track
  Proceedings}. OpenReview.net.

\bibitem[{Allen{-}Zhu and Li(2021)}]{DBLP:conf/focs/Allen-ZhuL21}
Zeyuan Allen{-}Zhu and Yuanzhi Li. 2021.
\newblock \href {https://doi.org/10.1109/FOCS52979.2021.00098} {Feature
  purification: How adversarial training performs robust deep learning}.
\newblock In \emph{62nd {IEEE} Annual Symposium on Foundations of Computer
  Science, {FOCS} 2021, Denver, CO, USA, February 7-10, 2022}, pages 977--988.
  {IEEE}.

\bibitem[{Alzantot et~al.(2018)Alzantot, Sharma, Elgohary, Ho, Srivastava, and
  Chang}]{alzantot-etal-2018-generating}
Moustafa Alzantot, Yash Sharma, Ahmed Elgohary, Bo-Jhang Ho, Mani Srivastava,
  and Kai-Wei Chang. 2018.
\newblock \href {https://doi.org/10.18653/v1/D18-1316} {Generating natural
  language adversarial examples}.
\newblock In \emph{Proceedings of the 2018 Conference on Empirical Methods in
  Natural Language Processing}, pages 2890--2896, Brussels, Belgium.
  Association for Computational Linguistics.

\bibitem[{Bao et~al.(2023)Bao, Zheng, Ding, Zhang, and
  Tao}]{bao-etal-2023-casn}
Rong Bao, Rui Zheng, Liang Ding, Qi~Zhang, and Dacheng Tao. 2023.
\newblock \href {https://doi.org/10.18653/v1/2023.acl-long.40}
  {{CASN}:class-aware score network for textual adversarial detection}.
\newblock In \emph{Proceedings of the 61st Annual Meeting of the Association
  for Computational Linguistics (Volume 1: Long Papers)}, pages 671--687,
  Toronto, Canada. Association for Computational Linguistics.

\bibitem[{Chen et~al.(2023)Chen, Ye, Zu, Xu, Zheng, Peng, Zhou, Gui, Zhang, and
  Huang}]{chen2023robust}
Xuanting Chen, Junjie Ye, Can Zu, Nuo Xu, Rui Zheng, Minlong Peng, Jie Zhou,
  Tao Gui, Qi~Zhang, and Xuanjing Huang. 2023.
\newblock How robust is gpt-3.5 to predecessors? a comprehensive study on
  language understanding tasks.
\newblock \emph{arXiv preprint arXiv:2303.00293}.

\bibitem[{Devlin et~al.(2019)Devlin, Chang, Lee, and
  Toutanova}]{devlin-etal-2019-bert}
Jacob Devlin, Ming-Wei Chang, Kenton Lee, and Kristina Toutanova. 2019.
\newblock \href {https://doi.org/10.18653/v1/N19-1423} {{BERT}: Pre-training of
  deep bidirectional transformers for language understanding}.
\newblock In \emph{Proceedings of the 2019 Conference of the North {A}merican
  Chapter of the Association for Computational Linguistics: Human Language
  Technologies, Volume 1 (Long and Short Papers)}, pages 4171--4186,
  Minneapolis, Minnesota. Association for Computational Linguistics.

\bibitem[{Ebrahimi et~al.(2018)Ebrahimi, Rao, Lowd, and
  Dou}]{ebrahimi-etal-2018-hotflip}
Javid Ebrahimi, Anyi Rao, Daniel Lowd, and Dejing Dou. 2018.
\newblock \href {https://doi.org/10.18653/v1/P18-2006} {{H}ot{F}lip: White-box
  adversarial examples for text classification}.
\newblock In \emph{Proceedings of the 56th Annual Meeting of the Association
  for Computational Linguistics (Volume 2: Short Papers)}, pages 31--36,
  Melbourne, Australia. Association for Computational Linguistics.

\bibitem[{Fischer and Alemi(2020)}]{DBLP:journals/entropy/FischerA20}
Ian Fischer and Alexander~A. Alemi. 2020.
\newblock \href {https://doi.org/10.3390/e22101081} {{CEB} improves model
  robustness}.
\newblock \emph{Entropy}, 22(10):1081.

\bibitem[{Gao et~al.(2018)Gao, Lanchantin, Soffa, and
  Qi}]{DBLP:conf/sp/GaoLSQ18}
Ji~Gao, Jack Lanchantin, Mary~Lou Soffa, and Yanjun Qi. 2018.
\newblock \href {https://doi.org/10.1109/SPW.2018.00016} {Black-box generation
  of adversarial text sequences to evade deep learning classifiers}.
\newblock In \emph{2018 {IEEE} Security and Privacy Workshops, {SP} Workshops
  2018, San Francisco, CA, USA, May 24, 2018}, pages 50--56. {IEEE} Computer
  Society.

\bibitem[{Garg and Ramakrishnan(2020)}]{garg-ramakrishnan-2020-bae}
Siddhant Garg and Goutham Ramakrishnan. 2020.
\newblock \href {https://doi.org/10.18653/v1/2020.emnlp-main.498} {{BAE}:
  {BERT}-based adversarial examples for text classification}.
\newblock In \emph{Proceedings of the 2020 Conference on Empirical Methods in
  Natural Language Processing (EMNLP)}, pages 6174--6181, Online. Association
  for Computational Linguistics.

\bibitem[{Goodfellow et~al.(2015)Goodfellow, Shlens, and
  Szegedy}]{DBLP:journals/corr/GoodfellowSS14}
Ian~J. Goodfellow, Jonathon Shlens, and Christian Szegedy. 2015.
\newblock \href {http://arxiv.org/abs/1412.6572} {Explaining and harnessing
  adversarial examples}.
\newblock In \emph{3rd International Conference on Learning Representations,
  {ICLR} 2015, San Diego, CA, USA, May 7-9, 2015, Conference Track
  Proceedings}.

\bibitem[{Gretton et~al.(2005)Gretton, Bousquet, Smola, and
  Sch{\"{o}}lkopf}]{DBLP:conf/alt/GrettonBSS05}
Arthur Gretton, Olivier Bousquet, Alexander~J. Smola, and Bernhard
  Sch{\"{o}}lkopf. 2005.
\newblock \href {https://doi.org/10.1007/11564089\_7} {Measuring statistical
  dependence with hilbert-schmidt norms}.
\newblock In \emph{Algorithmic Learning Theory, 16th International Conference,
  {ALT} 2005, Singapore, October 8-11, 2005, Proceedings}, volume 3734 of
  \emph{Lecture Notes in Computer Science}, pages 63--77. Springer.

\bibitem[{Gretton et~al.(2007)Gretton, Fukumizu, Teo, Song, Sch{\"{o}}lkopf,
  and Smola}]{DBLP:conf/nips/GrettonFTSSS07}
Arthur Gretton, Kenji Fukumizu, Choon~Hui Teo, Le~Song, Bernhard
  Sch{\"{o}}lkopf, and Alexander~J. Smola. 2007.
\newblock \href
  {https://proceedings.neurips.cc/paper/2007/hash/d5cfead94f5350c12c322b5b664544c1-Abstract.html}
  {A kernel statistical test of independence}.
\newblock In \emph{Advances in Neural Information Processing Systems 20,
  Proceedings of the Twenty-First Annual Conference on Neural Information
  Processing Systems, Vancouver, British Columbia, Canada, December 3-6, 2007},
  pages 585--592. Curran Associates, Inc.

\bibitem[{Horn and Johnson(2012)}]{horn2012matrix}
Roger~A Horn and Charles~R Johnson. 2012.
\newblock \emph{Matrix analysis}.
\newblock Cambridge university press.

\bibitem[{Ilyas et~al.(2019)Ilyas, Santurkar, Tsipras, Engstrom, Tran, and
  Madry}]{DBLP:conf/nips/IlyasSTETM19}
Andrew Ilyas, Shibani Santurkar, Dimitris Tsipras, Logan Engstrom, Brandon
  Tran, and Aleksander Madry. 2019.
\newblock \href
  {https://proceedings.neurips.cc/paper/2019/hash/e2c420d928d4bf8ce0ff2ec19b371514-Abstract.html}
  {Adversarial examples are not bugs, they are features}.
\newblock In \emph{Advances in Neural Information Processing Systems 32: Annual
  Conference on Neural Information Processing Systems 2019, NeurIPS 2019,
  December 8-14, 2019, Vancouver, BC, Canada}, pages 125--136.

\bibitem[{Iyyer et~al.(2018)Iyyer, Wieting, Gimpel, and
  Zettlemoyer}]{iyyer-etal-2018-adversarial}
Mohit Iyyer, John Wieting, Kevin Gimpel, and Luke Zettlemoyer. 2018.
\newblock \href {https://doi.org/10.18653/v1/N18-1170} {Adversarial example
  generation with syntactically controlled paraphrase networks}.
\newblock In \emph{Proceedings of the 2018 Conference of the North {A}merican
  Chapter of the Association for Computational Linguistics: Human Language
  Technologies, Volume 1 (Long Papers)}, pages 1875--1885, New Orleans,
  Louisiana. Association for Computational Linguistics.

\bibitem[{Jin et~al.(2020)Jin, Jin, Zhou, and
  Szolovits}]{DBLP:conf/aaai/JinJZS20}
Di~Jin, Zhijing Jin, Joey~Tianyi Zhou, and Peter Szolovits. 2020.
\newblock \href {https://ojs.aaai.org/index.php/AAAI/article/view/6311} {Is
  {BERT} really robust? {A} strong baseline for natural language attack on text
  classification and entailment}.
\newblock In \emph{The Thirty-Fourth {AAAI} Conference on Artificial
  Intelligence, {AAAI} 2020, The Thirty-Second Innovative Applications of
  Artificial Intelligence Conference, {IAAI} 2020, The Tenth {AAAI} Symposium
  on Educational Advances in Artificial Intelligence, {EAAI} 2020, New York,
  NY, USA, February 7-12, 2020}, pages 8018--8025. {AAAI} Press.

\bibitem[{Jolliffe and Cadima(2016)}]{jolliffe2016principal}
Ian~T Jolliffe and Jorge Cadima. 2016.
\newblock Principal component analysis: a review and recent developments.
\newblock \emph{Philosophical Transactions of the Royal Society A:
  Mathematical, Physical and Engineering Sciences}, 374(2065):20150202.

\bibitem[{Kim et~al.(2021)Kim, Lee, and Ro}]{DBLP:conf/nips/KimLR21}
Junho Kim, Byung{-}Kwan Lee, and Yong~Man Ro. 2021.
\newblock \href
  {https://proceedings.neurips.cc/paper/2021/hash/8e5e15c4e6d09c8333a17843461041a9-Abstract.html}
  {Distilling robust and non-robust features in adversarial examples by
  information bottleneck}.
\newblock In \emph{Advances in Neural Information Processing Systems 34: Annual
  Conference on Neural Information Processing Systems 2021, NeurIPS 2021,
  December 6-14, 2021, virtual}, pages 17148--17159.

\bibitem[{Kornblith et~al.(2019)Kornblith, Norouzi, Lee, and
  Hinton}]{DBLP:conf/icml/Kornblith0LH19}
Simon Kornblith, Mohammad Norouzi, Honglak Lee, and Geoffrey~E. Hinton. 2019.
\newblock \href {http://proceedings.mlr.press/v97/kornblith19a.html}
  {Similarity of neural network representations revisited}.
\newblock In \emph{Proceedings of the 36th International Conference on Machine
  Learning, {ICML} 2019, 9-15 June 2019, Long Beach, California, {USA}},
  volume~97 of \emph{Proceedings of Machine Learning Research}, pages
  3519--3529. {PMLR}.

\bibitem[{Kramer(1991)}]{kramer1991nonlinear}
Mark~A Kramer. 1991.
\newblock Nonlinear principal component analysis using autoassociative neural
  networks.
\newblock \emph{AIChE journal}, 37(2):233--243.

\bibitem[{LeCun et~al.(2015)LeCun, Bengio, and Hinton}]{lecun2015deep}
Yann LeCun, Yoshua Bengio, and Geoffrey Hinton. 2015.
\newblock Deep learning.
\newblock \emph{nature}, 521(7553):436--444.

\bibitem[{Li et~al.(2019)Li, Ji, Du, Li, and Wang}]{DBLP:conf/ndss/LiJDLW19}
Jinfeng Li, Shouling Ji, Tianyu Du, Bo~Li, and Ting Wang. 2019.
\newblock \href
  {https://www.ndss-symposium.org/ndss-paper/textbugger-generating-adversarial-text-against-real-world-applications/}
  {Textbugger: Generating adversarial text against real-world applications}.
\newblock In \emph{26th Annual Network and Distributed System Security
  Symposium, {NDSS} 2019, San Diego, California, USA, February 24-27, 2019}.
  The Internet Society.

\bibitem[{Li et~al.(2020{\natexlab{a}})Li, Ma, Guo, Xue, and
  Qiu}]{li-etal-2020-bert-attack}
Linyang Li, Ruotian Ma, Qipeng Guo, Xiangyang Xue, and Xipeng Qiu.
  2020{\natexlab{a}}.
\newblock \href {https://doi.org/10.18653/v1/2020.emnlp-main.500}
  {{BERT}-{ATTACK}: Adversarial attack against {BERT} using {BERT}}.
\newblock In \emph{Proceedings of the 2020 Conference on Empirical Methods in
  Natural Language Processing (EMNLP)}, pages 6193--6202, Online. Association
  for Computational Linguistics.

\bibitem[{Li et~al.(2020{\natexlab{b}})Li, Ma, Guo, Xue, and
  Qiu}]{DBLP:conf/emnlp/LiMGXQ20}
Linyang Li, Ruotian Ma, Qipeng Guo, Xiangyang Xue, and Xipeng Qiu.
  2020{\natexlab{b}}.
\newblock \href {https://doi.org/10.18653/v1/2020.emnlp-main.500}
  {{BERT-ATTACK:} adversarial attack against {BERT} using {BERT}}.
\newblock In \emph{Proceedings of the 2020 Conference on Empirical Methods in
  Natural Language Processing, {EMNLP} 2020, Online, November 16-20, 2020},
  pages 6193--6202. Association for Computational Linguistics.

\bibitem[{Lin et~al.(2021)Lin, Zou, and Ding}]{lin-etal-2021-using}
Jieyu Lin, Jiajie Zou, and Nai Ding. 2021.
\newblock \href {https://doi.org/10.18653/v1/2021.acl-short.43} {Using
  adversarial attacks to reveal the statistical bias in machine reading
  comprehension models}.
\newblock In \emph{Proceedings of the 59th Annual Meeting of the Association
  for Computational Linguistics and the 11th International Joint Conference on
  Natural Language Processing (Volume 2: Short Papers)}, pages 333--342,
  Online. Association for Computational Linguistics.

\bibitem[{Liu et~al.(2022)Liu, Zheng, Rong, Liu, Liu, Cheng, Qiao, Gui, Zhang,
  and Huang}]{liu-etal-2022-flooding}
Qin Liu, Rui Zheng, Bao Rong, Jingyi Liu, ZhiHua Liu, Zhanzhan Cheng, Liang
  Qiao, Tao Gui, Qi~Zhang, and Xuanjing Huang. 2022.
\newblock \href {https://doi.org/10.18653/v1/2022.acl-long.386} {Flooding-{X}:
  Improving {BERT}{'}s resistance to adversarial attacks via loss-restricted
  fine-tuning}.
\newblock In \emph{Proceedings of the 60th Annual Meeting of the Association
  for Computational Linguistics (Volume 1: Long Papers)}, pages 5634--5644,
  Dublin, Ireland. Association for Computational Linguistics.

\bibitem[{Liu et~al.(2019)Liu, Ott, Goyal, Du, Joshi, Chen, Levy, Lewis,
  Zettlemoyer, and Stoyanov}]{DBLP:journals/corr/abs-1907-11692}
Yinhan Liu, Myle Ott, Naman Goyal, Jingfei Du, Mandar Joshi, Danqi Chen, Omer
  Levy, Mike Lewis, Luke Zettlemoyer, and Veselin Stoyanov. 2019.
\newblock \href {http://arxiv.org/abs/1907.11692} {Roberta: {A} robustly
  optimized {BERT} pretraining approach}.
\newblock \emph{CoRR}, abs/1907.11692.

\bibitem[{Maas et~al.(2011)Maas, Daly, Pham, Huang, Ng, and
  Potts}]{maas-etal-2011-learning}
Andrew~L. Maas, Raymond~E. Daly, Peter~T. Pham, Dan Huang, Andrew~Y. Ng, and
  Christopher Potts. 2011.
\newblock \href {https://aclanthology.org/P11-1015} {Learning word vectors for
  sentiment analysis}.
\newblock In \emph{Proceedings of the 49th Annual Meeting of the Association
  for Computational Linguistics: Human Language Technologies}, pages 142--150,
  Portland, Oregon, USA. Association for Computational Linguistics.

\bibitem[{Madry et~al.(2018)Madry, Makelov, Schmidt, Tsipras, and
  Vladu}]{DBLP:conf/iclr/MadryMSTV18}
Aleksander Madry, Aleksandar Makelov, Ludwig Schmidt, Dimitris Tsipras, and
  Adrian Vladu. 2018.
\newblock \href {https://openreview.net/forum?id=rJzIBfZAb} {Towards deep
  learning models resistant to adversarial attacks}.
\newblock In \emph{6th International Conference on Learning Representations,
  {ICLR} 2018, Vancouver, BC, Canada, April 30 - May 3, 2018, Conference Track
  Proceedings}. OpenReview.net.

\bibitem[{Maheshwary et~al.(2021)Maheshwary, Maheshwary, and
  Pudi}]{DBLP:conf/aaai/MaheshwaryMP21}
Rishabh Maheshwary, Saket Maheshwary, and Vikram Pudi. 2021.
\newblock \href {https://ojs.aaai.org/index.php/AAAI/article/view/17595}
  {Generating natural language attacks in a hard label black box setting}.
\newblock In \emph{Thirty-Fifth {AAAI} Conference on Artificial Intelligence,
  {AAAI} 2021, Thirty-Third Conference on Innovative Applications of Artificial
  Intelligence, {IAAI} 2021, The Eleventh Symposium on Educational Advances in
  Artificial Intelligence, {EAAI} 2021, Virtual Event, February 2-9, 2021},
  pages 13525--13533. {AAAI} Press.

\bibitem[{Quadrianto et~al.(2019)Quadrianto, Sharmanska, and
  Thomas}]{DBLP:conf/cvpr/QuadriantoST19}
Novi Quadrianto, Viktoriia Sharmanska, and Oliver Thomas. 2019.
\newblock \href {https://doi.org/10.1109/CVPR.2019.00842} {Discovering fair
  representations in the data domain}.
\newblock In \emph{{IEEE} Conference on Computer Vision and Pattern
  Recognition, {CVPR} 2019, Long Beach, CA, USA, June 16-20, 2019}, pages
  8227--8236. Computer Vision Foundation / {IEEE}.

\bibitem[{Ren et~al.(2019)Ren, Deng, He, and Che}]{ren-etal-2019-generating}
Shuhuai Ren, Yihe Deng, Kun He, and Wanxiang Che. 2019.
\newblock \href {https://doi.org/10.18653/v1/P19-1103} {Generating natural
  language adversarial examples through probability weighted word saliency}.
\newblock In \emph{Proceedings of the 57th Annual Meeting of the Association
  for Computational Linguistics}, pages 1085--1097, Florence, Italy.
  Association for Computational Linguistics.

\bibitem[{Socher et~al.(2013)Socher, Perelygin, Wu, Chuang, Manning, Ng, and
  Potts}]{socher-etal-2013-recursive}
Richard Socher, Alex Perelygin, Jean Wu, Jason Chuang, Christopher~D. Manning,
  Andrew Ng, and Christopher Potts. 2013.
\newblock \href {https://aclanthology.org/D13-1170} {Recursive deep models for
  semantic compositionality over a sentiment treebank}.
\newblock In \emph{Proceedings of the 2013 Conference on Empirical Methods in
  Natural Language Processing}, pages 1631--1642, Seattle, Washington, USA.
  Association for Computational Linguistics.

\bibitem[{Song et~al.(2012)Song, Smola, Gretton, Bedo, and
  Borgwardt}]{DBLP:journals/jmlr/SongSGBB12}
Le~Song, Alexander~J. Smola, Arthur Gretton, Justin Bedo, and Karsten~M.
  Borgwardt. 2012.
\newblock \href {https://doi.org/10.5555/2503308.2343691} {Feature selection
  via dependence maximization}.
\newblock \emph{J. Mach. Learn. Res.}, 13:1393--1434.

\bibitem[{Szegedy et~al.(2014)Szegedy, Zaremba, Sutskever, Bruna, Erhan,
  Goodfellow, and Fergus}]{DBLP:journals/corr/SzegedyZSBEGF13}
Christian Szegedy, Wojciech Zaremba, Ilya Sutskever, Joan Bruna, Dumitru Erhan,
  Ian~J. Goodfellow, and Rob Fergus. 2014.
\newblock \href {http://arxiv.org/abs/1312.6199} {Intriguing properties of
  neural networks}.
\newblock In \emph{2nd International Conference on Learning Representations,
  {ICLR} 2014, Banff, AB, Canada, April 14-16, 2014, Conference Track
  Proceedings}.

\bibitem[{Tanay and Griffin(2016)}]{DBLP:journals/corr/TanayG16}
Thomas Tanay and Lewis~D. Griffin. 2016.
\newblock \href {http://arxiv.org/abs/1608.07690} {A boundary tilting
  persepective on the phenomenon of adversarial examples}.
\newblock \emph{CoRR}, abs/1608.07690.

\bibitem[{Tsipras et~al.(2019)Tsipras, Santurkar, Engstrom, Turner, and
  Madry}]{DBLP:conf/iclr/TsiprasSETM19}
Dimitris Tsipras, Shibani Santurkar, Logan Engstrom, Alexander Turner, and
  Aleksander Madry. 2019.
\newblock \href {https://openreview.net/forum?id=SyxAb30cY7} {Robustness may be
  at odds with accuracy}.
\newblock In \emph{7th International Conference on Learning Representations,
  {ICLR} 2019, New Orleans, LA, USA, May 6-9, 2019}. OpenReview.net.

\bibitem[{Vaswani et~al.(2018)Vaswani, Bouwmans, Javed, and
  Narayanamurthy}]{DBLP:journals/spm/VaswaniBJN18}
Namrata Vaswani, Thierry Bouwmans, Sajid Javed, and Praneeth Narayanamurthy.
  2018.
\newblock \href {https://doi.org/10.1109/MSP.2018.2826566} {Robust subspace
  learning: Robust pca, robust subspace tracking, and robust subspace
  recovery}.
\newblock \emph{{IEEE} Signal Process. Mag.}, 35(4):32--55.

\bibitem[{Wallace et~al.(2019)Wallace, Rodriguez, Feng, Yamada, and
  Boyd-Graber}]{wallace-etal-2019-trick}
Eric Wallace, Pedro Rodriguez, Shi Feng, Ikuya Yamada, and Jordan Boyd-Graber.
  2019.
\newblock \href {https://doi.org/10.1162/tacl_a_00279} {Trick me if you can:
  Human-in-the-loop generation of adversarial examples for question answering}.
\newblock \emph{Transactions of the Association for Computational Linguistics},
  7:387--401.

\bibitem[{Wang et~al.(2021{\natexlab{a}})Wang, Wang, Cheng, Gan, Jia, Li, and
  Liu}]{DBLP:conf/iclr/WangWCGJLL21}
Boxin Wang, Shuohang Wang, Yu~Cheng, Zhe Gan, Ruoxi Jia, Bo~Li, and Jingjing
  Liu. 2021{\natexlab{a}}.
\newblock \href {https://openreview.net/forum?id=hpH98mK5Puk} {Infobert:
  Improving robustness of language models from an information theoretic
  perspective}.
\newblock In \emph{9th International Conference on Learning Representations,
  {ICLR} 2021, Virtual Event, Austria, May 3-7, 2021}. OpenReview.net.

\bibitem[{Wang et~al.(2020)Wang, Wang, Qin, Packer, Li, Chen, Beutel, and
  Chi}]{wang-etal-2020-cat}
Tianlu Wang, Xuezhi Wang, Yao Qin, Ben Packer, Kang Li, Jilin Chen, Alex
  Beutel, and Ed~Chi. 2020.
\newblock \href {https://doi.org/10.18653/v1/2020.emnlp-main.417} {{CAT}-gen:
  Improving robustness in {NLP} models via controlled adversarial text
  generation}.
\newblock In \emph{Proceedings of the 2020 Conference on Empirical Methods in
  Natural Language Processing (EMNLP)}, pages 5141--5146, Online. Association
  for Computational Linguistics.

\bibitem[{Wang et~al.(2021{\natexlab{b}})Wang, Liu, Gui, Zhang, Zou, Zhou, Ye,
  Zhang, Zheng, Pang, Wu, Li, Zhang, Ma, Fei, Cai, Zhao, Hu, Yan, Tan, Hu,
  Bian, Liu, Qin, Zhu, Xing, Fu, Zhang, Peng, Zheng, Zhou, Wei, Qiu, and
  Huang}]{wang-etal-2021-textflint}
Xiao Wang, Qin Liu, Tao Gui, Qi~Zhang, Yicheng Zou, Xin Zhou, Jiacheng Ye,
  Yongxin Zhang, Rui Zheng, Zexiong Pang, Qinzhuo Wu, Zhengyan Li, Chong Zhang,
  Ruotian Ma, Zichu Fei, Ruijian Cai, Jun Zhao, Xingwu Hu, Zhiheng Yan, Yiding
  Tan, Yuan Hu, Qiyuan Bian, Zhihua Liu, Shan Qin, Bolin Zhu, Xiaoyu Xing,
  Jinlan Fu, Yue Zhang, Minlong Peng, Xiaoqing Zheng, Yaqian Zhou, Zhongyu Wei,
  Xipeng Qiu, and Xuanjing Huang. 2021{\natexlab{b}}.
\newblock \href {https://doi.org/10.18653/v1/2021.acl-demo.41} {{T}ext{F}lint:
  Unified multilingual robustness evaluation toolkit for natural language
  processing}.
\newblock In \emph{Proceedings of the 59th Annual Meeting of the Association
  for Computational Linguistics and the 11th International Joint Conference on
  Natural Language Processing: System Demonstrations}, pages 347--355, Online.
  Association for Computational Linguistics.

\bibitem[{Wolf et~al.(2020)Wolf, Debut, Sanh, Chaumond, Delangue, Moi, Cistac,
  Rault, Louf, Funtowicz, Davison, Shleifer, von Platen, Ma, Jernite, Plu, Xu,
  Le~Scao, Gugger, Drame, Lhoest, and Rush}]{wolf-etal-2020-transformers}
Thomas Wolf, Lysandre Debut, Victor Sanh, Julien Chaumond, Clement Delangue,
  Anthony Moi, Pierric Cistac, Tim Rault, Remi Louf, Morgan Funtowicz, Joe
  Davison, Sam Shleifer, Patrick von Platen, Clara Ma, Yacine Jernite, Julien
  Plu, Canwen Xu, Teven Le~Scao, Sylvain Gugger, Mariama Drame, Quentin Lhoest,
  and Alexander Rush. 2020.
\newblock \href {https://doi.org/10.18653/v1/2020.emnlp-demos.6} {Transformers:
  State-of-the-art natural language processing}.
\newblock In \emph{Proceedings of the 2020 Conference on Empirical Methods in
  Natural Language Processing: System Demonstrations}, pages 38--45, Online.
  Association for Computational Linguistics.

\bibitem[{Zang et~al.(2020)Zang, Qi, Yang, Liu, Zhang, Liu, and
  Sun}]{zang-etal-2020-word}
Yuan Zang, Fanchao Qi, Chenghao Yang, Zhiyuan Liu, Meng Zhang, Qun Liu, and
  Maosong Sun. 2020.
\newblock \href {https://doi.org/10.18653/v1/2020.acl-main.540} {Word-level
  textual adversarial attacking as combinatorial optimization}.
\newblock In \emph{Proceedings of the 58th Annual Meeting of the Association
  for Computational Linguistics}, pages 6066--6080, Online. Association for
  Computational Linguistics.

\bibitem[{Zhang et~al.(2020)Zhang, Sheng, Alhazmi, and
  Li}]{DBLP:journals/tist/ZhangSAL20}
Wei~Emma Zhang, Quan~Z. Sheng, Ahoud Alhazmi, and Chenliang Li. 2020.
\newblock \href {https://doi.org/10.1145/3374217} {Adversarial attacks on
  deep-learning models in natural language processing: {A} survey}.
\newblock \emph{{ACM} Trans. Intell. Syst. Technol.}, 11(3):24:1--24:41.

\bibitem[{Zhang et~al.(2015)Zhang, Zhao, and LeCun}]{DBLP:conf/nips/ZhangZL15}
Xiang Zhang, Junbo~Jake Zhao, and Yann LeCun. 2015.
\newblock \href
  {https://proceedings.neurips.cc/paper/2015/hash/250cf8b51c773f3f8dc8b4be867a9a02-Abstract.html}
  {Character-level convolutional networks for text classification}.
\newblock In \emph{Advances in Neural Information Processing Systems 28: Annual
  Conference on Neural Information Processing Systems 2015, December 7-12,
  2015, Montreal, Quebec, Canada}, pages 649--657.

\bibitem[{Zhang et~al.(2021)Zhang, Zhang, Chen, and
  He}]{zhang-etal-2021-crafting}
Xinze Zhang, Junzhe Zhang, Zhenhua Chen, and Kun He. 2021.
\newblock \href {https://doi.org/10.18653/v1/2021.acl-long.153} {Crafting
  adversarial examples for neural machine translation}.
\newblock In \emph{Proceedings of the 59th Annual Meeting of the Association
  for Computational Linguistics and the 11th International Joint Conference on
  Natural Language Processing (Volume 1: Long Papers)}, pages 1967--1977,
  Online. Association for Computational Linguistics.

\bibitem[{Zheng et~al.(2023{\natexlab{a}})Zheng, Dou, Zhou, Liu, Gui, Zhang,
  Wei, Huang, and Zhang}]{zheng-etal-2023-detecting}
Rui Zheng, Shihan Dou, Yuhao Zhou, Qin Liu, Tao Gui, Qi~Zhang, Zhongyu Wei,
  Xuanjing Huang, and Menghan Zhang. 2023{\natexlab{a}}.
\newblock \href {https://doi.org/10.18653/v1/2023.findings-acl.717} {Detecting
  adversarial samples through sharpness of loss landscape}.
\newblock In \emph{Findings of the Association for Computational Linguistics:
  ACL 2023}, pages 11282--11298, Toronto, Canada. Association for Computational
  Linguistics.

\bibitem[{Zheng et~al.(2022)Zheng, Rong, Zhou, Liang, Wang, Wu, Gui, Zhang, and
  Huang}]{zheng-etal-2022-robust}
Rui Zheng, Bao Rong, Yuhao Zhou, Di~Liang, Sirui Wang, Wei Wu, Tao Gui,
  Qi~Zhang, and Xuanjing Huang. 2022.
\newblock \href {https://doi.org/10.18653/v1/2022.acl-long.157} {Robust lottery
  tickets for pre-trained language models}.
\newblock In \emph{Proceedings of the 60th Annual Meeting of the Association
  for Computational Linguistics (Volume 1: Long Papers)}, pages 2211--2224,
  Dublin, Ireland. Association for Computational Linguistics.

\bibitem[{Zheng et~al.(2023{\natexlab{b}})Zheng, Xi, Liu, Lai, Gui, Zhang,
  Huang, Ma, Shan, and Ge}]{zheng-etal-2023-characterizing}
Rui Zheng, Zhiheng Xi, Qin Liu, Wenbin Lai, Tao Gui, Qi~Zhang, Xuanjing Huang,
  Jin Ma, Ying Shan, and Weifeng Ge. 2023{\natexlab{b}}.
\newblock \href {https://doi.org/10.18653/v1/2023.findings-acl.146}
  {Characterizing the impacts of instances on robustness}.
\newblock In \emph{Findings of the Association for Computational Linguistics:
  ACL 2023}, pages 2314--2332, Toronto, Canada. Association for Computational
  Linguistics.

\bibitem[{Zheng et~al.(2023{\natexlab{c}})Zheng, Xi, Liu, Lai, Gui, Zhang,
  Huang, Ma, Shan, and Ge}]{DBLP:conf/acl/ZhengXLLGZHMSG23}
Rui Zheng, Zhiheng Xi, Qin Liu, Wenbin Lai, Tao Gui, Qi~Zhang, Xuanjing Huang,
  Jin Ma, Ying Shan, and Weifeng Ge. 2023{\natexlab{c}}.
\newblock \href {https://doi.org/10.18653/v1/2023.findings-acl.146}
  {Characterizing the impacts of instances on robustness}.
\newblock In \emph{Findings of the Association for Computational Linguistics:
  {ACL} 2023, Toronto, Canada, July 9-14, 2023}, pages 2314--2332. Association
  for Computational Linguistics.

\bibitem[{Zhu et~al.(2020)Zhu, Cheng, Gan, Sun, Goldstein, and
  Liu}]{DBLP:conf/iclr/ZhuCGSGL20}
Chen Zhu, Yu~Cheng, Zhe Gan, Siqi Sun, Tom Goldstein, and Jingjing Liu. 2020.
\newblock \href {https://openreview.net/forum?id=BygzbyHFvB} {Freelb: Enhanced
  adversarial training for natural language understanding}.
\newblock In \emph{8th International Conference on Learning Representations,
  {ICLR} 2020, Addis Ababa, Ethiopia, April 26-30, 2020}. OpenReview.net.

\end{thebibliography}
